%% file: acl_latex.tex
\pdfoutput=1
\documentclass[11pt]{article}

\usepackage[preprint]{acl}

\usepackage{tabularx}
\usepackage{multirow}
\usepackage{bm}
\usepackage{array}
\usepackage{CJKutf8}
\usepackage{makecell}
\usepackage{graphicx}
\usepackage{amsmath}
\usepackage{algorithm, algpseudocode}
\usepackage{hyperref}
\usepackage{multirow}
\usepackage{booktabs}
\usepackage{xcolor}
\usepackage{amsfonts}
\usepackage{pifont} % For check marks and crosses
\algrenewcommand\algorithmicrequire{\textbf{Input:}}
\algrenewcommand\algorithmicensure{\textbf{Output:}}

\usepackage{times}
\usepackage{latexsym}

\usepackage[T1]{fontenc}
\usepackage[utf8]{inputenc}

\usepackage{microtype}

\usepackage{inconsolata}

\usepackage{graphicx}
\usepackage{makecell}
\usepackage{tabularx}
\usepackage{longtable}
\usepackage{tabularx}

\newcommand*\colourcheck[1]{%
  \expandafter\newcommand\csname #1check\endcsname{\textcolor{#1}{\ding{52}}}%
}

\colourcheck{green}

\newcommand*\colourcross[1]{%
  \expandafter\newcommand\csname #1cross\endcsname{\textcolor{#1}{\ding{56}}}%
}

\colourcross{red}

\title{QualBench: Benchmarking Chinese LLMs with Localized Professional Qualifications for Vertical Domain Evaluation}

\author{Mengze Hong$^{1,2}$, Wailing Ng$^{1}$, Chen Jason Zhang$^{1}$, Di Jiang$^{1}$\thanks{Corresponding Author}\\ 
$^{1}$Hong Kong Polytechnic University, $^{2}$AI Group, WeBank Co., Ltd\\
}

\begin{document}
\maketitle
\begin{abstract}

The rapid advancement of Chinese LLMs underscores the need for vertical-domain evaluations to ensure reliable applications. However, existing benchmarks often lack domain coverage and provide limited insights into the Chinese working context. Leveraging qualification exams as a unified framework for expertise evaluation, we introduce \textbf{QualBench}\footnote{Data and code are publicly available at \url{https://github.com/mengze-hong/QualBench}}, the first multi-domain Chinese QA benchmark dedicated to localized assessment of Chinese LLMs. The dataset includes over 17,000 questions across six vertical domains, drawn from 24 Chinese qualifications to align with national policies and professional standards. Results reveal an interesting pattern of Chinese LLMs consistently surpassing non-Chinese models, with the Qwen2.5 model outperforming the more advanced GPT-4o, emphasizing the value of localized domain knowledge in meeting qualification requirements. The average accuracy of 53.98\% reveals the current gaps in domain coverage within model capabilities. Furthermore, we identify performance degradation caused by LLM crowdsourcing, assess data contamination, and illustrate the effectiveness of prompt engineering and model fine-tuning, suggesting opportunities for future improvements through multi-domain RAG and Federated Learning.
\end{abstract}

\section{Introduction}
Large Language Models (LLMs) trained on Chinese corpora have recently gained attention for their improved Chinese text generation and understanding \cite{guo2025deepseek, yang2024qwen2, yang-etal-2025-fraud}. While researchers increasingly claim their superiority over humans in various NLP tasks \cite{doi:10.1073/pnas.2305016120}, existing benchmark evaluations mainly focus on language capabilities and often neglect the assessment of domain knowledge, offering limited quantitative evidence of their effectiveness in downstream applications \cite{10.5555/3491440.3491941, NEURIPS2023_7a92bcde}. This gap is especially pronounced given the rapid advancement of both Chinese and non-Chinese LLMs, raising the question: Which model should be used?

\begin{CJK}{UTF8}{gbsn}

\begin{table}[!t]
    \centering
    \resizebox{\columnwidth}{!}{
    \begin{tabular}{c|l}
        \toprule
        General & \thead[l]{货币的首要或基本功能是什么？\\What is the primary function of money?\\A. 交易工具 (Medium of Exchange) \\ B. 储存方式 (Store of Value)\\C. 付款方式 (Means of Payment) \\ D. 价值衡量标准 (Measure of Value)} \\ \midrule
        Localized   & \thead[l]{人民币是指\_\_\_\_依法发行的货币。\\The renminbi refers to currency issued by\_\_\_\_ according to law.\\A. 中国人民银行 (People's Bank of China) \\ B. 中国银行 (Bank of China)\\C. 中国工商银行 (Industrial and Commercial Bank of China) \\ D. 银行业监督管理委员会 (China Banking Regulatory Commission)} \\
        \bottomrule
        \end{tabular}}
    \caption{Example of general and localized knowledge evaluation questions in the banking domain.}
    \label{tab:localization}
    \vspace{-1em}
\end{table}

\end{CJK}

A notable trend in knowledge evaluation is the use of qualification examinations, which provide a unified framework for assessing job readiness and domain expertise \cite{oxenham2024education}, and are increasingly valued for their fairness and transparency \cite{li2024lexeval}. While many qualification tests, such as the Chinese College Entrance Examination (Gaokao) \cite{zong-qiu-2024-gaokao} and the National Civil Servants Examination \cite{10.5555/3491440.3491941}, have been used to assess LLMs, numerous domain-specific evaluations remain underutilized. These vertical-domain qualifications are highly diverse and offer undeniable advantages for evaluating LLMs’ domain knowledge in a localized Chinese context, as illustrated in Table \ref{tab:localization}.

\input{floats/tab_datasets_summary}

In this paper, we introduce the \underline{Qual}ification \underline{Bench}mark (\textbf{QualBench}), providing comprehensive vertical domain evaluations based on professional qualification exams in China. We detail the data construction process and, through extensive experiments, demonstrate the challenges posed by the dataset. The dataset serves as an essential safeguard for reliable LLM deployment and application in China. In summary, the contributions are:

\begin{enumerate}
\item We summarize the limitations of existing Chinese benchmark datasets and highlight the importance of incorporating qualification exams into LLM evaluation for a comprehensive understanding of downstream applicability.

\item We release the first multi-domain Chinese benchmark dataset grounded in 24 qualification examinations across six vertical domains. This dataset emphasizes localization and provides practical insights into LLM capabilities within the Chinese working context, offering human-aligned expertise assessment.

\item We present several interesting findings: (1) Chinese LLMs consistently outperform non-Chinese models regardless of model size; (2) LLM crowdsourcing results in performance degradation, with a single strong model outperforming the collaborative efforts of multiple LLMs; and (3) knowledge augmentation through prompting and fine-tuning has a greater impact on non-Chinese LLMs. These observations underscore the importance of localized domain knowledge for Chinese qualifications and highlight the need to expand single-model knowledge coverage through RAG augmentation and Federated Learning.

\end{enumerate}

\section{Preliminaries}

\paragraph{Benchmark with Qualification Examinations.} Qualification examinations are rigorously verified by domain experts before public release, offering invaluable benefits to constructing realistic and reliable benchmark datasets \cite{yang2024crag, zhong-etal-2024-agieval}. These exams serve as crucial gateways for certifying professionals in specific job roles, making them ideal tools for evaluating LLMs prior to deployment in real-world tasks \cite{katz2024gpt}. Moreover, existing domain-specific LLMs predominantly focus on areas like medicine, law, and finance, representing only a small subset of vertical domains \cite{singhal2023large, 10.1145/3641289}. Incorporating qualification exams introduces great diversity to the dataset construction, leading towards a better understanding of LLMs' capabilities across different domains while offering human-aligned expertise evaluation scores for easy interpretation \cite{ling2023domain, cui2025compositionalarchitectureregretlarge}.

\paragraph{Localization and Domain Coverage.} Localization is a crucial factor for identifying suitable domain experts within specific contexts. For instance, a lawyer trained in the United States cannot legally practice in China due to differences in legal systems. Existing qualification exam–based benchmarks (Table~\ref{tab:overview}) often overlook distinctions between Chinese and international contexts. This lack of localization is reflected in both the choice of data sources that involve generic qualifications \cite{zhu-etal-2024-benchmarking}  and in evaluation results that frequently conclude state-of-the-art performance with English LLMs such as GPT and LLaMA \cite{zhang2023evaluating, li2024lexeval}. Additionally, existing evaluations often focus on single-domain evaluation, which limits complexity and enables domain-specific models, such as FinGPT \cite{luukkonen-etal-2023-fingpt}, to consistently dominate their benchmarks. Such narrow coverage offers little insight into LLM capability and their real-world applicability, which often requires cross-domain knowledge \cite{hong2025expandingchatbotknowledgecustomer, liu2025bridge}.

\section{Dataset Construction}
In this paper, we argue that a well-constructed benchmark for evaluating Chinese LLMs should feature localized Chinese contexts and cover diverse vertical domains for sufficient complexity.

\subsection{Data Sources}

We concentrate on six vertical domains, each featuring at least three professional qualification examinations to ensure dataset reliability. In total, 24 qualification exams are included, each encompassing up to 10 years of collected examination papers (see Table \ref{tab:exams}). Optical Character Recognition is used to extract the questions, answers, and explanations (when applicable) from the PDF documents, resulting in 31,841 QA pairs in the initial dataset.

\subsection{Data Preprocessing}

We exclude questions that rely on non-textual information to avoid inconsistent evaluation results stemming from varying capabilities in visual understanding. Repeated questions are removed from the dataset using both similarity matching and human screening. Notably, many similar questions recur across qualification exams within the same domain over different years, especially those related to government policy and industry regulations, reflecting the dataset’s enduring relevance \cite{yang2024crag}. The dataset is further validated by two domain experts per field, tasked with assessing question relevance and QA pair completeness.

\begin{CJK}{UTF8}{gbsn}
\begin{table}[!t]
\small
\centering
\resizebox{\columnwidth}{!}{
\begin{tabular}{lc}
\toprule
\textbf{Category} & \textbf{Number of Questions} \\ \midrule
Production Safety (安全生产) & 6520 \\
Fire Safety (消防安全) & 3401 \\
Civil Engineering (建筑工程) & 1978 \\
Economics and Finance (经济金融) & 2377 \\
Oil and Gas (石油天然气) & 1604 \\
Banking and Insurance (银行保险) & 1436 \\ \midrule
\textbf{Total} & 17,316 \\ 
\bottomrule
\end{tabular}}
\caption{Dataset statistics: number of questions across different domains.}
\label{tab: stats}
\end{table}
\end{CJK}

\subsection{Dataset Statistics}
The dataset consists of 17,316 knowledge-driven questions, including 9,538 single-choice, 3,710 multiple-choice questions with up to eight answer choices, and 4,068 True/False questions. The average length for each question type is 83, 91, and 48, respectively. As shown in Table \ref{tab: stats}, the dataset is imbalanced with a skew towards ``Production Safety'' and ``Fire Safety''. These two domains have been excessively overlooked in existing benchmark datasets but hold significant importance due to their localized characteristics and strong correlation with government policy. Thus, more questions are included to fill in the gap. The sample questions and demonstration of each question type are presented in Appendix \ref{sec:demo}.

\input{floats/tab_main_results}

\begin{table}[!t]
\centering
\small
\resizebox{\columnwidth}{!}{
\begin{tabular}{lcccc}
\toprule
\multirow{2}{*}{\textbf{Model}} & \multicolumn{2}{c}{\textbf{Small Variant} {\tiny (7B)}} & \multicolumn{2}{c}{\textbf{Large Variant} {\tiny (13B / 14B)}} \\
\cmidrule(lr){2-3} \cmidrule(lr){4-5}
& \textbf{Acc (\%)} & \textbf{F1 (\%)} & \textbf{Acc (\%)} & \textbf{F1 (\%)} \\
\midrule
Baichuan2 & \textbf{48.64} & \textbf{49.27} & \textbf{49.73} & \textbf{51.55 }\\
LLama   & 30.45 & 31.28 & 37.65 & 38.17 \\ 
\bottomrule
\end{tabular}
}
\caption{Performance comparison between small and large variants of Chinese and non-Chinese LLMs.}

\label{tab:size_compare}
\end{table}

\section{Experimental Setup}

To thoroughly assess the capabilities of Chinese LLMs in vertical domain QAs, we evaluate five widely deployed and accessible models: ChatGLM3, Qwen2.5, Baichuan2, Hunyuan-v2, and Deepseek-v2. Given the dataset's focus on localization, we also compare with non-Chinese LLMs, including Mistral-7B, LLaMa-7B, GPT-3.5, and GPT4o, to reveal the importance of Chinese knowledge for achieving satisfactory performance. GPT models are accessed via the OpenAI API, while the others are deployed from Hugging Face with inference run on a single A100 GPU.

All models are evaluated in a one-shot setting to standardize the output format. Each model is prompted to provide both an answer and an explanation to facilitate result interpretation. We set hyperparameters with a temperature of 0.5, top\_p of 0.9, and a maximum token limit of 1024 to control the diversity and creativity of generated answers. Inference was repeated five times, and we report the average accuracy and F1 score for robustness.

\section{Results and Discussion}

\subsection{Main Results}
As shown in Table~\ref{tab:overall comparison}, Qwen2.5 consistently achieves superior performance across different domains, benefiting from its extensive pre-trained knowledge in the Chinese context. The second and third-best models, GPT-4o and GPT-3.5, have substantially more parameters but still underperform compared to the smaller 7B model. A general pattern emerges: \textbf{Chinese LLMs consistently outperform non-Chinese models on QualBench}. This observation is further strengthened by comparing models of different sizes (see Table~\ref{tab:size_compare}), which shows that while larger LLMs generally perform better, smaller Chinese models (e.g., Baichuan2-7B, 48.64\% accuracy) can still outperform larger non-Chinese models (e.g., LLaMA-14B, 37.65\%) due to their stronger localized expertise. These results underscore the importance of native Chinese knowledge for achieving better performance on QualBench, offering a comprehensive evaluation of LLM applicability in Chinese work environments. The findings also contrast with most existing benchmarks, where larger models typically outperform smaller ones regardless of origin~\cite{ni2025surveylargelanguagemodel}.

Table~\ref{tab: question type} presents performance across each question type, revealing Qwen2.5’s notable superiority and its substantial advantage over GPT-4o in both single-choice and multiple-choice questions. The error analysis in Appendix~\ref{sec:error} further identifies five distinct error patterns in the dataset, revealing specific limitations in model capabilities and highlighting directions for future improvements.

\subsection{LLM Crowdsourcing}

Recent research highlights the potential of LLM crowds to mitigate biases inherent in relying on a single model \cite{zhao-etal-2022-lmturk, hong2025llm, he-etal-2024-annollm}. However, other work cautions that different LLMs may often produce highly similar responses, offering limited improvement at high cost \cite{hong2025dialin}. In our benchmark, we observed a more pronounced issue of \textbf{performance degradation}: both voting-based aggregation methods combining five LLMs failed to surpass the performance of a single Qwen model, primarily due to substantial performance gaps among the models, which introduced undesirable noise into the final outcome. These findings suggest that a single robust LLM can potentially serve as a more effective and cost-efficient choice for knowledge-intensive tasks, while also highlighting the need for further advancements in aggregation techniques to better leverage the diverse outputs of multiple LLMs.

\begin{table}[!t]
\centering
\small
\resizebox{\columnwidth}{!}{
\begin{tabular}{lcc}
\toprule
\textbf{Model} & \textbf{Accuracy (\%) - Original} & \textbf{Accuracy (\%) - Shuffled} \\
\midrule
Qwen2.5-7b     & 74.78 & 74.45 \\
Qwen2.5-14b    & 79.04 & 78.52 \\
Mistral-7b  & 35.17 & 32.73 \\
GPT-3.5     & 52.24 & 50.93 \\
\bottomrule
\end{tabular}}
\caption{Performance comparison between original and shuffled datasets.}
\label{tab:data-contamination}
\end{table}

\subsection{Ablation Studies}

We present ablation studies to assess key factors affecting evaluation robustness and LLM performance, detailing the characteristics of QualBench.

\paragraph{Data Contamination and Robustness.}
To address potential data contamination that could bias LLM performance comparisons on our proposed benchmark \cite{sainz-etal-2023-nlp, golchin2025data}, two ablation studies are conducted to ensure robustness. First, we implement an \textbf{\textit{answer-shuffling strategy}} that randomizes the answer order for both single- and multiple-choice questions to create a shuffled dataset. Unlike altering question type or length \cite{zhao2025largelanguagemodelsbadly}, this method preserves the original contextual information while disrupting the question–answer mapping, enabling assessment of potential memorization. As shown in Table~\ref{tab:data-contamination}, the input perturbation resulted in only minor accuracy changes, likely due to generation randomness (temp = 0.5), suggesting that LLMs rely primarily on the provided question context rather than retrieving memorized responses from their training data \cite{schwarzschild2024rethinking}.

\begin{table}[!t]
\centering
\small
\resizebox{\columnwidth}{!}{
\begin{tabular}{cccccc}
\toprule
\textbf{$n$} & \textbf{Edit Dist.} & \textbf{ROUGE-L F1} & \textbf{Match $\ge$ 80\%} & \textbf{Acc (\%)} \\
\midrule
1020 & 20.35\% & 28.40\% & 5.10\% & 29.61\% \\
\bottomrule
\end{tabular}}
\caption{Performance of question completion with Qwen2.5-7B.}
\vspace{-1em}
\label{tab:question-completion}
\end{table}

Second, we evaluated the best-performing Qwen2.5-7B model in a \textbf{\textit{question completion task}}, where 1,020 questions were randomly sampled. For each question, half of the text was removed from the end, and the model was instructed to reconstruct the complete question and then provide an answer. This test rigorously assessed the extent to which the LLM could recall the original questions. Results in Table~\ref{tab:question-completion} show extremely low similarity between the generated and original questions, along with significantly lower answer accuracy, suggesting that the questions from QualBench are largely absent from the model’s prior knowledge. Overall, results show that QualBench has minimal overlap with LLM training data, suggesting little to no impact from data contamination on benchmark performance.

\paragraph{Geographic Contextual Prompting.}

Given the strong geographical characteristics of our dataset, we conduct an ablation study to evaluate the effect of geographic contextual prompting on LLM performance \cite{brown2020language}. The prompt enhancements include: (1) a geographically grounded role description (e.g., Chinese expert profile) and (2) in‑context learning (ICL) with domain‑specific demonstrations, highlighting the role of contextual information in enabling LLMs to function as domain experts in localized settings. As shown in Table~\ref{tab:prompt}, the Chinese LLM (qwen-7b) exhibits only marginal performance gains from these enhancements, likely due to its substantial inherent local knowledge. In contrast, English LLMs demonstrate greater improvements when provided with localized expertise; however, they still lag behind the Chinese model, emphasizing the advantage conferred by pre-trained Chinese local knowledge.

\begin{table}[!t]
\centering 
\resizebox{\columnwidth}{!}{ \small
\begin{tabular}{lcccc}
\toprule
\textbf{Model} & \textbf{Original (\%)} & \textbf{Role (\%)} & \textbf{ICL (\%)} & \textbf{Gain (\%)} \\
\midrule
Qwen2.5-7b     & 75.26 & 75.63 & 76.08 & +0.82 \\
Mistral-7b  & 43.03 & 46.55 & 48.09 & +5.06 \\
Llama-7b    & 30.45 & 32.76 & 38.92 & +8.47 \\
\bottomrule
\end{tabular}}
\caption{Performance (accuracy) comparison of different prompt configurations.}
\label{tab:prompt}
\end{table}

\paragraph{LLM Fine-tuning.} 

We evaluate the DISC-FinLLM model \cite{chen2023disc}, a domain-specific LLM based on Baichuan2-13B fine-tuned on Chinese finance corpora, on the ``Economics and Finance'' subset. We further fine-tuned DISC-FinLLM on 30\% of the subset\footnote{Fine-tuning was performed using LoRA with a learning rate of $1\text{e}^{-5}$, trained for 5 epochs with the AdamW optimizer.} to demonstrate the value of the proposed dataset for training future vertical-domain LLMs. Results in Table~\ref{tab:finetune_results} show that the DISC-FinLLM (40.19\%) underperformed the base Baichuan2-13B model (41.97\%), likely due to limited coverage of in-depth qualification exam knowledge and noise from prior training on other datasets. However, fine-tuning DISC-FinLLM on QualBench improved accuracy to 44.43\%, highlighting the dataset’s unique value in enhancing domain-specific expertise and underscoring its complexity for evaluating LLMs.

\begin{table}[!t]
\centering
\resizebox{\columnwidth}{!}{
\begin{tabular}{llc}
\toprule
\textbf{Model} & \textbf{Accuracy (\%)} \\
\midrule
Baichuan2-13b (vanilla) & 41.97 \\
\midrule
DISC-FinLLM (base) & 40.19 \\
\midrule
DISC-FinLLM (fine-tuned on QualBench) & \textbf{44.43} \\
\bottomrule
\end{tabular}}
\caption{Performance of domain-specific DISC-FinLLM and its fine-tuned variant on QualBench.}
\vspace{-1.5em}
\label{tab:finetune_results}
\end{table}

\section{Discussion}

The current best performance of 75.25\% represents a marginal pass of the qualification, underscoring the need for further improvement. In response, we present and analyze two promising subroutines:

\paragraph{Retrieval Augmented Generation (RAG).} While incorporating external knowledge through RAG can effectively boost LLM performance \cite{mao-etal-2024-rag, wang2024domainrag, zhang2025erarag}, constructing a comprehensive RAG knowledge base for the proposed dataset is challenging due to the presence of multiple domains. In practice, passing all 24 qualification exams would require extensive domain-specific study materials, including textbooks, industry standards, and regulatory guidelines. The volume and diversity of this knowledge make relevant retrieval costly and computation-intensive. This calls for more efficient approaches, such as cross-domain knowledge graphs \cite{liu2022cross} and online resource integration \cite{arslan2024survey}, to link localized, domain-specific contexts into a cohesive multi-domain knowledge base.

\paragraph{LLM Training with Federated Learning.} Training vertical LLMs within the Chinese context presents valuable opportunities for practical deployment. However, most existing domain-specific LLMs struggle with highly specialized tasks due to the data-hungry issue that limits knowledge coverage \cite{10.1145/3298981, villalobos2024position}. To address this challenge, Federated Learning can be harnessed to effectively utilize private domain data from companies and government agencies \cite{jiang2019federated}, enabling privacy-preserving access to training data \cite{fan2023fate, kuang2024federatedscope}.

\section{Conclusion}
In this work, we present QualBench, the first multi-domain LLM benchmark drawn from 24 Chinese qualification exams spanning six vertical domains. Our evaluation shows that Chinese LLMs consistently outperform non-Chinese models, regardless of model size. We also expose the limitations of LLM crowdsourcing and suggest promising avenues for future research. Overall, QualBench underscores the value of qualification exams as a rich resource for human-aligned expertise assessment in LLMs, motivating future efforts to incorporate more diverse formats and broader coverage.

\section*{Limitations}
While the presented benchmark offers a realistic vertical-domain expertise assessment, it has several limitations. First, the dataset exhibits domain imbalance, with areas such as Production Safety and Fire Safety intentionally over-represented to enable deeper domain-specific evaluation. This skews the overall performance profile and may provide an incomplete picture of model capabilities in less-represented domains. Second, the benchmark focuses exclusively on multiple-choice and true/false questions, leaving open-ended formats unexplored. Third, all questions are text-only for consistency, omitting image and audio modalities. Future benchmarks could reveal richer insights by integrating open-ended formats with figure-aided questions, assessing the more complex domain-specific reasoning and problem-solving abilities required in real-world applications.

\section*{Ethical Consideration}

The ethical collection and responsible use of the presented dataset are our top priorities. All questions are sourced from publicly accessible Chinese qualification examinations, ensuring compliance with relevant regulations. The dataset is designed exclusively for academic research and knowledge evaluation, with commercial use strictly prohibited. We have rigorously verified the authenticity and accuracy of all questions to establish a reliable vertical-domain benchmark for LLMs.

\section*{Acknowledgments}

We express our gratitude to Prof. Hongyu Lin for his valuable feedback and support during the development of this benchmark dataset and the preparation of the manuscript. We also extend our sincere appreciation to the anonymous reviewers for their insightful and constructive comments, which significantly enhanced the quality of this work.

This paper is partially supported by several ongoing projects led or coordinated by Prof. Zhang Chen, including including P0045948 and P0046453 (industry donations from Accel Group Holding Limited and Minshang Creative Technology Holdings Limited), P0046701 and P0046703 (PolyU internal research funding), P0048887 (Innovation and Technology Fund - ITSP, ITS/028/22FP),  P0051906 (RGC Early Career Scheme, 25600624), and P0054482 (Two Square Capital Limited donation).

\bibliography{custom}

\appendix

\input{floats/tab_question_type_performance}
\input{floats/tab_exam_sources}

\section{Error Analysis}
\label{sec:error}

To characterize the error patterns of LLM evaluation on QualBench dataset, we analyzed 1,200 incorrect predictions from the test outputs. Each was manually inspected to identify question characteristics. From this process, we derived five distinct error categories, each reflecting specific LLM limitations. These categories, along with sample questions, are presented in Table~\ref{tab:error-analysis}.

\section{Demonstration of Domain Coverage and Question Types}
\label{sec:demo}
Table~\ref{tab:exams} details the Chinese qualification examinations used as data sources. Tables~\ref{tab:PS} to \ref{tab:BI} present sample QAs from each of the six domains, with two questions for each of the three question types. This demonstration highlights both the breadth of domain coverage and the variety of question formats, ensuring a comprehensive evaluation of LLM performance across different areas of expertise.

\input{floats/tab_error_analysis}

\input{floats/tab_domain_demos}

\end{document}

%% file: floats/tab_datasets_summary.tex
\begin{table*}[ht]
    \centering
    \resizebox{\textwidth}{!}{
    \begin{tabular}{llcllcc}
        \toprule
        \textbf{Dataset} & \textbf{Source Qualification Exam} & \textbf{Size} & \textbf{Best Model} & \textbf{Vertical Domain} & \textbf{Localization}  & \textbf{Explainable} \\
        \midrule
        GAOKAO-Bench \cite{zhang2023evaluating} & Chinese College Entrance Examination (Gaokao) & 2811 & GPT-4 & \redcross & \greencheck & \redcross  \\
        LexEval \cite{li2024lexeval} & National Unified Legal Professional Qualification & 14,150 & GPT-4  & Legal & \greencheck & \redcross  \\
        MedBench \cite{cai2024medbench} & Medical Qualification Exams & 40,041 & GPT-4 & Medical & \redcross & \redcross \\
        CFLUE \cite{zhu-etal-2024-benchmarking} & Finance Qualification Exams & 38,636 & Qwen-72B & Finance & \redcross & \greencheck \\
        M3KE \cite{liu2023m3ke} & Entrance Exams of Different Education Levels & 20,477 & GPT-3.5 & \redcross & \greencheck & \redcross \\
        FinEval \cite{zhang2023fineval} & Finance Qualification Exams & 8,351 &GPT4o& Finance & \redcross & \redcross \\
        CMExam \cite{liu2023benchmarking} & Chinese National Medical Licensing Exam & 68,119 & GPT-4 & Medical & \redcross & \redcross \\
        LogiQA \cite{10.5555/3491440.3491941} & Civil Servants Exams of China & 8,678 & RoBERTa & \redcross & \greencheck & \greencheck \\
        % JEC-QA  \cite{zhong2020jec} & National Judicial Examination of China & 26,365 && Legal & \greencheck & \redcross \\
        \midrule
        \textbf{QualBench (ours)} & Multiple Sources & 17,316 & Qwen-7B & Multiple & \greencheck & \greencheck \\
        \bottomrule
    \end{tabular}}
    \caption{Overview of existing Chinese benchmark datasets constructed based on qualification exams.}
    \vspace{-1em}
    \label{tab:overview}
\end{table*}

%% file: floats/tab_main_results.tex
\begin{CJK}{UTF8}{gbsn}
\begin{table*}[!t]
\small
\centering
\resizebox{\textwidth}{!}{
\begin{tabular}{@{}llccccccccccccc@{}}
\toprule
& \multirow{3}{*}{\textbf{Model}} & \multicolumn{2}{c}{\textbf{Production Safety}} & \multicolumn{2}{c}{\textbf{Oil and Gas}} & \multicolumn{2}{c}{\textbf{Fire Safety}} & \multicolumn{2}{c}{\textbf{Civil Engineering}} & \multicolumn{2}{c}{\textbf{Economics and Finance}} & \multicolumn{2}{c}{\textbf{Banking and Insurance}} & \multirow{3}{*}{\textbf{Overall}} \\
& & \multicolumn{2}{c}{(安全生产)} & \multicolumn{2}{c}{(石油天然气)} & \multicolumn{2}{c}{(消防安全)} & \multicolumn{2}{c}{(建筑工程)} & \multicolumn{2}{c}{(经济金融)} & \multicolumn{2}{c}{(银行保险)} \\
\cmidrule(lr){3-4} \cmidrule(lr){5-6} \cmidrule(lr){7-8} \cmidrule(lr){9-10} \cmidrule(lr){11-12} \cmidrule(lr){13-14}
& & Acc (\%) & F1 (\%) & Acc (\%) & F1 (\%) & Acc (\%) & F1 (\%) & Acc (\%) & F1 (\%) & Acc (\%) & F1 (\%) & Acc (\%) & F1 (\%)  \\ \midrule
\multirow{5}{*}{\textbf{Chinese}} & Chatglm3-6b-chat & 41.42 & 45.40 & 48.88 & 45.36 & 53.53 & 54.52 & 39.45 & 39.91 & 38.72 & 40.31 & 43.75 & 47.56 & 43.62\% \\
& Baichuan2-7b-chat & 49.69 & 50.71 & 60.85 & 61.68 & 52.11 & 52.95 & 40.44 & 42.06 & 42.13 & 43.61 & 48.50 & 49.93 & 48.64\% \\
& \textbf{Qwen2.5-7b-instruct} & \textbf{76.98} & \textbf{78.39} & 71.82 & 75.13 & \textbf{78.56} & \textbf{79.10} & \textbf{63.05} & \textbf{64.59} & \textbf{77.52} & \textbf{77.81} & \textbf{82.53} & \textbf{83.20} & \textbf{75.26\%} \\
& Hunyuan-7b & 50.11 & 53.90 & 50.62 & 58.57 & 55.77 & 57.79 & 43.56 & 46.48 & 50.40 & 52.68 & 55.91 & 59.77 & 50.64\% \\
& Deepseek-v2-lite-chat & 48.72 & 52.74 & 57.92 & 53.87 & 58.61 & 59.28 & 45.66 & 45.84 & 49.94 & 50.04 & 59.05 & 60.58 & 51.76\% \\ \midrule
\multirow{4}{*}{\textbf{Non-Chinese}} & Mistral-7b-instruct & 43.25 & 43.54 & 51.43 & 50.69 & 48.07 & 49.13 & 36.83 & 35.72 & 38.21 & 38.66 & 41.58 & 41.63 & 43.03\% \\
& LLama-7b & 32.20 & 32.39 & 33.92 & 33.20 & 35.68 & 36.19 & 22.65 & 23.43 & 27.71 & 27.16 & 26.62 & 27.37 & 30.45\% \\
& GPT-3.5 & 58.93 & 59.76 & 69.33 & 70.43 & 53.32 & 55.13 & 47.41 & 49.39 & 54.28 & 54.34 & 62.54 & 63.23 & 56.96\% \\
& GPT-4o & 63.50 & 64.37 & \textbf{73.75} & \textbf{74.84} & 58.25 & 60.16 & 53.23 & 54.84 & 58.20 & 58.61 & 66.46 & 67.16 & 61.61\% \\ \midrule
\multirow{2}{*}{\textbf{Aggregation}} & majority voting & 56.98 & 59.61 & 60.97 & 66.69 & 66.24 & 66.35 & 53.39 & 53.78 & 59.51 & 59.95 & 67.58 & 68.57 & 59.56\% \\
& weighted majority voting & 61.79 & 64.45 & 63.28 & 69.03 & 69.51 & 70.06 & 57.54 & 58.36 & 64.95 & 65.27 & 73.72 & 74.59 & 63.98\% \\
\bottomrule
\end{tabular}}
\caption{Main evaluation results on QualBench. The best results are \textbf{bolded}.}
\label{tab:overall comparison}
\end{table*}
\end{CJK}

%% file: floats/tab_question_type_performance.tex
\begin{table*}[!t]
% \resizebox{\columnwidth}{!}{
\centering
\small
\begin{tabular}{llcccccc}
\toprule
& \multirow{2}{*}{\textbf{Model}} & \multicolumn{2}{c}{\textbf{Single Choice}} & \multicolumn{2}{c}{\textbf{Multiple Choice}} & \multicolumn{2}{c}{\textbf{True/False}} \\
\cmidrule(lr){3-4} \cmidrule(lr){5-6} \cmidrule(lr){7-8}
& & Acc & F1 & Acc & F1 & Acc & F1 \\
\midrule
\multirow{5}{*}{\textbf{Chinese}} & Chatglm3-6b-chat & 0.4773 & 0.4800 & 0.1676 & 0.2112 & 0.5530 & 0.5552 \\
& Baichuan2-7b-chat & 0.5061 & 0.5072 & 0.1453 & 0.1986 & 0.7030 & 0.7036 \\
& Qwen2.5-7b-instruct & \textbf{0.8089} & \textbf{0.8094} & \textbf{0.6117} & \textbf{0.6262} & \textbf{0.7409} & 0.7470 \\
& Hunyuan-7b & 0.5643 & 0.5817 & 0.2534 & 0.3179 & 0.5764 & 0.5994 \\
& DeepSeek-v2-lite-chat & 0.5567 & 0.5589 & 0.3362 & 0.3500 & 0.5726 & 0.5798 \\ 
\midrule
\multirow{4}{*}{\textbf{Non-Chinese}} & Mistral-7b-instruct-v0.3 & 0.4373 & 0.4392 & 0.1511 & 0.1845 & 0.6268 & 0.6310 \\
& LLama-7b & 0.2803 & 0.2852 & 0.0881 & 0.1070 & 0.5193 & 0.5188 \\
& GPT-3.5 & 0.6023 & 0.6062 & 0.3338 & 0.3591 & 0.6794 & 0.7034 \\
& GPT-4o & 0.6513 & 0.6575 & 0.3539 & 0.4039 & 0.7407 & \textbf{0.7516} \\
\midrule
\multirow{2}{*}{\textbf{Aggregation}} & majority vote & 0.6524 & 0.6511 & 0.4078 & 0.4402 & 0.6181 & 0.5995 \\
& weighted majority vote & 0.6961 & 0.6954 & 0.5121 & 0.5284 & 0.6181 & 0.5995 \\
% & Markov network & & & & & & \\
\bottomrule
\end{tabular}
% }
\caption{Performance across different question types.}
\label{tab: question type}
\end{table*}

%% file: floats/tab_exam_sources.tex
\begin{table*}[!t]
\centering
\begin{CJK}{UTF8}{gbsn}
\resizebox{\textwidth}{!}{
\begin{tabular}{lll}
\hline
\toprule
\makecell[c]{\textbf{Domain}} & \makecell[c]{\textbf{Qualification Name}} & \makecell[c]{\textbf{Qualification Name (English)}} \\
\midrule                                                    
Production Safety              & 中级注册安全工程师《安全生产》                                                & Intermediate Registered Safety Engineer "Safety Production"            \\ \midrule
Production Safety              & 建筑施工项目负责人B证                                                                 & Construction Project Manager B Certificate                                              \\ \hline
Production Safety              & 煤矿安全生产和管理能力                                                                           & Coal Mine Safety Production and Management Ability                                                             \\ \midrule
Production Safety              & 铁路安全试题                                                                                    & Railway Safety Exam Questions                                                                                  \\ \midrule
Civil Engineering              & 全国二级建造师《水利水电工程管理与实务》                                  &  National Level II Constructor "Water Conservancy and Hydropower Engineering Management and Practice" \\ \midrule
Civil Engineering              & 二建《管理》                                                                          & Second Builder "Management" Exam                        \\ \midrule
Civil Engineering              & 二级造价工程师《造价管理》                                                      & Level II Cost Engineer "Cost Management"                                             \\ \midrule
Civil Engineering              & 咨询工程师（投资）《宏观经济政策与发展规划》 & Consulting Engineer (Investment) "Macroeconomic Policy and Development Planning"\\ \midrule
Civil Engineering              & 注册土木工程师（岩土）《专业基础考试》题库   & Registered Civil Engineer (Geotechnical) "Professional Basic Exam"  \\ \midrule
Civil Engineering              & 注册土木工程师（道路工程）《专业考试》& Registered Civil Engineer (Road Engineering) "Professional Exam" \\ \midrule
Fire Safety                    & 火灾救援                                                                                        & Fire Rescue                                                                                            \\ \midrule
Fire Safety                    & 职业技能鉴定理论                                                                                & Occupational Skills Assessment Theory                                                                           \\ \midrule
Fire Safety                    & 设施操作员（初中高级）                                                                     & Facility Operator (Elementary, Intermediate, Advanced)                                   \\ \midrule
Oil and Gas                    & 天然气安全生产管理人员                                                                          & Natural Gas Safety Production Management Personnel                                                              \\ \midrule
Oil and Gas                    & 陆上石油天然气开采安全管理人员                                                            & Onshore Oil and Gas Extraction Safety Management Personnel                                     \\ \midrule
Economics and Finance          & 中级经济师《人力》                                                                    & Intermediate Economist "Human Resources"                                                     \\ \midrule
Economics and Finance          & 中级经济师《工商》                                                               & Intermediate Economist "Commerce"                                                  \\ \midrule
Economics and Finance          & 反假币                                                                                      & Counterfeit Currency                                                                                       \\ \midrule
Economics and Finance          & 经济师《经济基础知识（中级）》                                     & Economist "Economic Fundamentals (Intermediate)"    \\ \midrule
Economics and Finance          & 中级经济师金融知识                                                                              & Intermediate Economist Financial Knowledge                                                                                                                                                            \\ \midrule
Banking and Insurance          & 银行业专业人员职业资格考试& Banking Industry Professional Qualification Examination" \\ \midrule
Banking and Insurance          & 保险从业资格考试                                                                                            & Insurance Qualification Examination                                                                                                      \\ \midrule
Banking and Insurance          & 银行业法律法规与综合能力                                                                         & Banking Laws and Regulations and Comprehensive Ability                                                          \\ \midrule
Banking and Insurance          & 银行从业法律法规与综合能力                                                                   & Banking Professional Laws and Regulations and Comprehensive Ability                              \\ \bottomrule
\end{tabular}}
\caption{Overview of qualification examination sources included in QualBench dataset.}
\label{tab:exams}
\end{CJK}
\end{table*}

%% file: floats/tab_error_analysis.tex
\begin{CJK}{UTF8}{gbsn}
\onecolumn
\begin{longtable}{>{\raggedright\arraybackslash}m{0.35\textwidth}
                  >{\raggedright\arraybackslash}p{0.65\textwidth}}
\toprule
\textbf{Error Category and Description} & \textbf{Examples} \\
\midrule
\endfirsthead
\toprule
\textbf{Error Category and Description} & \textbf{Examples} \\
\midrule
\endhead

\bottomrule
\multicolumn{2}{@{}l@{}}{\vspace{0.1em}} \\[-0.1em]
\caption{Error categories with descriptions and demonstration of error questions.}
\label{tab:error-analysis-continued}
\endfoot

\bottomrule
\multicolumn{2}{@{}l@{}}{\vspace{0.1em}} \\[-0.1em]
\caption{Error categories with descriptions and demonstration of error questions.}
\label{tab:error-analysis}
\endlastfoot
\small
\textbf{Legal and Regulatory Comprehension.} Questions in this group emphasize the model’s understanding of legal norms, compliance rules, and administrative procedures specific to Chinese regulatory systems. They require interpreting statutory language, understanding obligations or restrictions, and distinguishing between subtly varied legal statements. Failure typically results from outdated or insufficient legal corpora, the model’s inability to disambiguate complex regulatory phrasing, as well as discrepancies in different regulatory systems, causing significant performance deviation between Chinese and non-Chinese LLMs. &
\tiny
\begin{tabularx}{\linewidth}{>{\raggedright\arraybackslash}X}
\toprule
\textbf{Question}: 依据《危险化学品安全管理条例》，下列剧毒化学品经营企业的行为中，正确是()。 \newline According to the Regulations on the Safety Management of Hazardous Chemicals, which of the following behaviors by enterprises operating highly toxic chemicals is correct? () \newline A．规定经营剧毒化学品销售记录的保存期限为1年 \newline B．规定经营剧毒化学品人员经过国家授权部门的专业培训合格后即可上岗 \newline C．规定经营剧毒化学品人员经过县级公安部门的专门培训合格后即可上岗 \newline D．向当地县级人民政府公安机关口头汇报购买的剧毒化学品数量和品种 \newline A. Stipulates that sales records of highly toxic chemicals must be retained for 1 year \newline B. Stipulates that personnel operating highly toxic chemicals may begin work after passing professional training conducted by a nationally authorized department \newline C. Stipulates that personnel operating highly toxic chemicals may begin work after passing special training conducted by the county-level public security department \newline D. Reports the quantity and type of highly toxic chemicals purchased to the county-level public security authority verbally \newline \textbf{Answer}: A \newline \textbf{LLM Results}: \newline chatglm3-6b-chat: D \quad qwen2.5-7b-instruct: B \quad baichuan2-7b-chat: B \newline hunyuan-7b: B \quad deepseek-v2-lite-chat: B \quad mistral-7b-instruct: D \newline LLama-7b: C \quad GPT-3.5: B \quad GPT-4o: B \\ \midrule
\textbf{Question}: 根据《中国国务院银行业监督管理机构行政复议办法》规定，下列有关行政复议的表述，正确的是()。 \newline According to the Measures for Administrative Reconsideration by the Banking Regulatory Authority of the State Council of China, which of the following statements about administrative reconsideration is correct? () \newline A. 复议机关依法复议后不得再提起行政诉讼 \newline B. 申请人申请行政复议，必须采用书面形式，不得口头申请 \newline C. 行政复议只审查具体行政行为是否合法 \newline D. 公民、法人或其他组织认为行政机关的具体行政行为侵犯其合法权益，可以向该行政机关申请复议 \newline A. After the reconsideration authority conducts a lawful review, administrative litigation may no longer be initiated \newline B. Applicants must submit administrative reconsideration requests in written form; oral applications are not allowed \newline C. Administrative reconsideration only examines whether a specific administrative act is lawful \newline D. Citizens, legal persons, or other organizations who believe that a specific administrative act has infringed their lawful rights and interests may apply for reconsideration to the same administrative authority \newline \textbf{Answer}: B \newline \textbf{LLM Results}: \newline chatglm3-6b-chat: D \quad qwen2.5-7b-instruct: A \quad baichuan2-7b-chat: D \newline hunyuan-7b: D \quad deepseek-v2-lite-chat: D \quad mistral-7b-instruct: D \newline LLama-7b: D \quad GPT-3.5: D \quad GPT-4o: D \\
\bottomrule
\end{tabularx} \\
\midrule
\small \textbf{Contextual Knowledge of National Systems.} This category involves questions requiring contextual understanding of China’s socio-political institutions, policy evolution, or economic infrastructure. Examples include classification of tax categories, stages of regulatory development, or administrative roles unique to China. Errors highlight challenges LLMs face with culturally and historically grounded knowledge, attributed to insufficient localization in model training, contributing significantly to performance deviation. &
\tiny
\begin{tabularx}{0.65\textwidth}{X}
\toprule
\textbf{Question}: 我国税收收入中的主体税种包括()。 \newline The primary types of taxes contributing to China's tax revenue include: () \newline A．所得税 \quad B．增值税 \quad C．资源税 \quad D．财产税 \quad E．消费税 \newline A. Income tax \quad B. Value-added tax \quad C. Resource tax \quad D. Property tax \newline E. Consumption tax \newline \textbf{Answer}: BE \newline \textbf{LLM Results}: \newline chatglm3-6b-chat: BDE \quad qwen2.5-7b-instruct: AB \quad baichuan2-7b-chat: ABE \newline hunyuan-7b: ABE \quad deepseek-v2-lite-chat: ABCE \quad mistral-7b-instruct: ABE \newline LLama-7b: ABCDE \quad GPT-3.5: ABCE \quad GPT-4o: AB \\ \midrule
\textbf{Question}: 我国银行监管在初步确立阶段的特点有()。 \newline Which of the following are characteristics of the initial establishment stage of China's banking regulatory system? () \newline A．监管职责配置的部门化，采取了功能化的监管组织架构 \newline B．监管运行机制的概念尚未提出，更缺少制度上的安排 \newline C．初步形成了机构监管的组织架构 \newline D．监管部门之间“各自为战”的现象较为突出 \newline E．监管治理得到完善，银行监管工作进入一个新的发展阶段 \newline A. The allocation of regulatory responsibilities was departmentalized, adopting a functional regulatory organizational structure \newline B. The concept of a regulatory operating mechanism had not yet been proposed, and there was a lack of institutional arrangements \newline C. A regulatory organizational structure based on institutions had begun to take shape \newline D. A relatively prominent phenomenon of “fighting their own battles” existed among regulatory departments \newline E. Regulatory governance had been improved, and banking regulation had entered a new stage of development \newline \textbf{Answer}: AB \newline \textbf{LLM Results}: \newline chatglm3-6b-chat: BCD \quad qwen2.5-7b-instruct: BCD \quad baichuan2-7b-chat: AD \newline hunyuan-7b: ABCDE \quad deepseek-v2-lite-chat: BD \quad mistral-7b-instruct: B \newline LLama-7b: CE \quad GPT-3.5: ABCD \quad GPT-4o: BCD \\
\bottomrule
\end{tabularx} \\
\small \textbf{Numerical Reasoning and Formula Application.} This category encompasses questions requiring arithmetic computation, quantitative estimation, or application of domain-specific formulas. Examples involve scenario-based problem solving or retrieval of correct analytical expressions. These questions assess a model’s ability to parse mathematical semantics, execute multi-step reasoning, and perform symbolic operations. &
\tiny
\begin{tabularx}{0.65\textwidth}{X}
\toprule
\textbf{Question}: 在有乙烷爆炸性危险的生产场所，对可能引起火灾的设备，可采用充氮气正压保护。假如乙醇不发生爆炸时氧的最高含量为11\%（体积比），空气中氧气占比为21\%，某设备内原有空气55L。为了防止该设备引起火灾或爆炸。采用充氮气泵的保护，氮气的需用量应不小于()。 \newline In production areas with the risk of ethane explosion, for equipment that may cause fire, nitrogen positive pressure protection can be used. Assume that the maximum oxygen content at which ethanol does not explode is 11\% (by volume), and the oxygen proportion in air is 21\%. There is originally 55L of air inside a certain piece of equipment. In order to prevent this equipment from causing fire or explosion, nitrogen is pumped in for protection. The required amount of nitrogen should be no less than: () \newline A．65L \quad B．60L \quad C．50L \quad D．55L \newline \textbf{Answer}: C \newline \textbf{LLM Results}: \newline chatglm3-6b-chat: B \quad qwen2.5-7b-instruct: A \quad baichuan2-7b-chat: B \newline hunyuan-7b: A \quad deepseek-v2-lite-chat: D \quad mistral-7b-instruct: A \newline LLama-7b: B \quad GPT-3.5: A \quad GPT-4o: B \\ \midrule
\textbf{Question}: 下列关于银行市值的说法，不正确的有()。 \newline Which of the following statements about bank market capitalization are incorrect? () \newline A．总市值等于发行总股份数乘以股票市价 \newline B．总市值等于发行总股份数乘以股票面值 \newline C．是衡量银行规模的重要综合性指标 \newline D．以H股为基准的市值（美元）＝（A股股价×A股股数＋H股股价×H股股数/港元对人民币汇率）/人民币对美元汇率 \newline E．以H股为基准的市值（美元）＝（A股股价×A股股数/人民币对港元汇率＋H股股价×H股股数）/港元对美元汇率 \newline A. Total market capitalization equals the total number of issued shares multiplied by the stock market price. \newline B. Total market capitalization equals the total number of issued shares multiplied by the stock par value. \newline C. It is an important comprehensive indicator for measuring the size of a bank. \newline D. Market capitalization (in USD) based based on H shares = (A-share price × number of A shares + H-share price × number of H shares / HKD to RMB exchange rate) / RMB to USD exchange rate. \newline E. Market capitalization (in USD) based on H shares = (A-share price × number of A shares / RMB to HKD exchange rate + H-share price × number of H shares) / HKD to USD exchange rate. \newline \textbf{Answer}: BD \newline \textbf{LLM Results}: \newline chatglm3-6b-chat: B \quad qwen2.5-7b-instruct: B \quad baichuan2-7b-chat: B \newline hunyuan-7b: ABCD \quad deepseek-v2-lite-chat: BDE \quad mistral-7b-instruct: B \newline LLama-7b: AE \quad GPT-3.5: BDE \quad GPT-4o: BE \\
\bottomrule
\end{tabularx} \\
\midrule
\small \textbf{Domain-Specific Inference.} Errors in this category arise from questions demanding deep, domain-specialized knowledge not readily inferable from general-purpose corpora. Successful resolution requires precise understanding of expert-level concepts, taxonomies, or operational practices within vertical fields. Failure suggests insufficient domain adaptation or inadequate exposure to fine-grained professional content during pretraining. &
\tiny
\begin{tabularx}{0.65\textwidth}{X}
\toprule
\textbf{Question}: 火灾探测器的工作原理是将烟雾、温度、火焰和燃烧气体等参量的变化通过敏感元件转化为电信号，传输到火灾报警控制器，不同种类的火灾探测器适用不同的场合。关于火灾探测器适用场合的说法，正确的是()。 \newline The working principle of Fire detectors is to convert changes in parameters such as smoke, temperature, flame, and combustion gases into electrical signals via sensitive components, which are then transmitted to the fire alarm control panel. Different types of fire detectors are suitable for different scenarios. Which of the following statements about the applicable scenarios of fire detectors is correct? () \newline A．感光探测适用于有易燃阶段的燃料火灾的场合 \newline B．红外火焰探测器适合于有大量烟雾存在的场合 \newline C．紫外火焰探测器特别适用于无机化合物燃烧的场合 \newline D．光电式感烟火灾探测器适用于发出黑烟的场合 \newline A. Photoelectric detection is suitable for flammable-stage fuel fires. \newline B. Infrared flame detectors are suitable for scenarios with a large amount of smoke present. \newline C. Ultraviolet flame detectors are particularly suitable for inorganic compound combustion scenarios. \newline D. Photoelectric smoke fire detectors are suitable for scenarios that emit black smoke. \newline \textbf{Answer}: B \newline \textbf{LLM Results}: \newline chatglm3-6b-chat: A \quad qwen2.5-7b-instruct: A \quad baichuan2-7b-chat: D \newline hunyuan-7b: C \quad deepseek-v2-lite-chat: D \quad mistral-7b-instruct: A \newline LLama-7b: A \quad GPT-3.5: D \quad GPT-4o: D \\ \midrule
\textbf{Question}: 下列关于贷款分类的说法中，错误的有()。 \newline Which of the following statements about loan classification are incorrect? () \newline A. 综合考虑了客户信用风险因素和债项交易损失因素 \newline B. 它实际上是根据预期损失对信贷资产进行评级 \newline C. 主要用于贷后管理，更多地体现为事后评价 \newline D. 通常仅考虑影响债项交易损失的特定风险因素，客户信用风险因素由客户评级完成 \newline E. 可同时用于贷前审批、贷后管理，是对债项风险的一种预先判断 \newline A. It comprehensively considers both the borrower's credit risk factors and the transaction loss factors of the obligation. \newline B. It essentially rates credit assets based on expected loss. \newline C. It is mainly used for post-loan management and more reflects post-event evaluation. \newline D. It usually considers only specific risk factors affecting transaction losses of the obligation, while the borrower's credit risk is handled through customer rating. \newline E. It can be used for both pre-loan approval and post-loan management, and serves as a forward-looking assessment of obligation risk. \newline \textbf{Answer}: DE \newline \textbf{LLM Results}: \newline chatglm3-6b-chat: BCD \quad qwen2.5-7b-instruct: CE \quad baichuan2-7b-chat: AD \newline hunyuan-7b: ABCD \quad deepseek-v2-lite-chat: BD \quad mistral-7b-instruct: C \newline LLama-7b: ABCE \quad GPT-3.5: D \quad GPT-4o: A \\
\bottomrule
\end{tabularx} \\
\small \textbf{Factual Detail Retrieval.} Questions that hinge on recalling exact factual details, such as numeric thresholds, legal deadlines, or procedural intervals, that cannot be inferred through reasoning or paraphrastic similarity. These items often occur sparsely in training data and require high-fidelity memorization. LLMs tend to hallucinate contextually plausible but incorrect answers, revealing a core weakness in factual grounding and precise retrieval from long-tail knowledge. &
\tiny
\begin{tabularx}{0.65\textwidth}{X}
\toprule
\textbf{Question}: 北京市《有限空间作业安全技术规范》规定，应急救援设备设施应根据同时开展有限空间作业点的数量进行配置，有多个作业点的，应在作业点()m范围内配置1套。 \newline
According to the Beijing Technical Specification for Safety in Confined Space Operations, emergency rescue equipment and facilities should be allocated based on the number of confined space operation sites being carried out simultaneously. If there are multiple operation sites, one set of equipment should be placed within a range of () meters from each site. \newline
A．100 \quad B．200 \quad C．400 \quad D．500 \newline
\textbf{Answer}: C \newline
\textbf{LLM Results}: \newline
chatglm3-6b-chat: B \quad qwen2.5-7b-instruct: A \quad baichuan2-7b-chat: D \newline
hunyuan-7b: B \quad deepseek-v2-lite-chat: B \quad mistral-7b-instruct: B \newline
LLama-7b: D \quad GPT-3.5: B \quad GPT-4o: B
\\ \midrule
\textbf{Question}: 商业银行应()披露其从事理财业务活动的有关信息。 \newline
Commercial banks should disclose information related to their wealth management business activities (). \newline
A. 每年 \quad B. 每半年 \quad C. 每月 \quad D. 每季度 \newline
A. Annually. \quad B. Semiannually. \quad C. Monthly. \quad D. Quarterly. \newline
\textbf{Answer}: B \newline
\textbf{LLM Results}: \newline
chatglm3-6b-chat: A \quad qwen2.5-7b-instruct: D \quad baichuan2-7b-chat: C \newline
hunyuan-7b: A \quad deepseek-v2-lite-chat: D \quad mistral-7b-instruct: D \newline
LLama-7b: C \quad GPT-3.5: D \quad GPT-4o: D
\\
\bottomrule
\end{tabularx} \\

\end{longtable}
\end{CJK}

%% file: floats/tab_domain_demos.tex
\begin{CJK}{UTF8}{gbsn}

\begin{table*}[ht]
\centering
\tiny
\renewcommand{\arraystretch}{1.2}  % optional: controls line spacing within cells
\begin{tabularx}{\textwidth}{
  >{\centering\arraybackslash}m{2cm}|
  >{\raggedright\arraybackslash}X|
  >{\raggedright\arraybackslash}X}
\toprule
\textbf{Category} & \makecell[c]{\textbf{Example 1}} & \makecell[c]{\textbf{Example 2}} \\
\midrule

\rule{0pt}{22ex}\shortstack{Production Safety\\(Multiple Choices)}
&
\vspace*{-12ex}
\textbf{Question}: 工作许可证制度要求，工作终结时，经()双方到现场交接验收。
\vspace{0.5em}\newline The work permit system requires that, upon completion of the work, on-site handover and acceptance shall be conducted by both: ()
\vspace{0.5em}\newline A．作业负责人
B．工作许可签发人
C．作业人员
D．政府主管部门
\vspace{0.5em}\newline A. The person in charge of the operation
B. The issuer of the work permit
C. The operating personnel
D. The competent government department
\vspace{0.5em}\newline\textbf{Answer}: AB
% \vspace{0.5em}\newline\textbf{Explanation}: /
&
\vspace*{-12ex}
\textbf{Question}: 主要负责人未履行法定安全生产管理职责而导致事故的，安全生产监督管理部门给予的罚款处罚说法正确的是()。
\vspace{0.5em}\newline In cases where the principal responsible person fails to perform the legally mandated duties for work safety management, resulting in an accident, which of the following statements about the fine imposed by the work safety supervision and administration department is correct? ()
\vspace{0.5em}\newline A．事故等级越高罚款金额越高
B．事故等级与罚款金额没有关系
C．发生特别重大事故处上一年年收入100\%的罚款
D．罚款金额以上一年收入为计算基数
\vspace{0.5em}\newline A. The higher the accident level, the higher the fine amount
B. The accident level has no relation to the fine amount
C. A particularly serious accident will result in a fine equal to 100\% of the person’s previous year's income
D. The fine amount is calculated based on the previous year's income
\vspace{0.5em}\newline\textbf{Answer}: ACD
% \vspace{0.5em}\newline\textbf{Explanation}: /
\\
\midrule
\rule{0pt}{24ex}\shortstack{Production Safety\\(Single Choice)}
&
\vspace*{-13ex}
\textbf{Question}: 生产经营单位的()必须按照国家有关规定经专门的安全作业培训，取得相应资格，方可上岗。
\vspace{0.5em}\newline Personnel engaged in () at a production and business operation unit must undergo specialized safety operation training in accordance with relevant national regulations, and obtain the corresponding qualifications before being permitted to start work.
\vspace{0.5em}\newline A．从业人员
\newline B．劳务派遣人员
\newline C．特种作业人员
\vspace{0.5em}\newline A. Employees
\newline B. Labor dispatch personnel
\newline C. Special operations personnel
\vspace{0.5em}\newline\textbf{Answer}: C
% \vspace{0.5em}\newline\textbf{Explanation}: /
&
\vspace*{-13ex}
\textbf{Question}: 《安全生产事故隐患排查治理暂行规定》中的安全生产事故隐患，是指生产经营单位违反()。
\vspace{0.5em}\newline In the Interim Provisions on the Investigation and Management of Potential Work Safety Accidents, a potential work safety accident refers to the violation by a production and business operation unit of: ()
\vspace{0.5em}\newline A．各种危险源
\newline B．物的危险状态、人的不安全行为和管理上的缺陷
\newline C．各类危险物品
\vspace{0.5em}\newline A. Various sources of danger
\newline B. Hazardous physical conditions, unsafe human behaviors, and management deficiencies
\newline C. Various hazardous substances
\vspace{0.5em}\newline\textbf{Answer}: B
% \vspace{0.5em}\newline\textbf{Explanation}: /
\\
\midrule
\rule{0pt}{12ex}\shortstack{Production Safety\\(True False)}
&
\vspace*{-7ex}
\textbf{Question}: 《安全生产法》规定，建设项目安全设施的设计人、设计单位应当对安全设施设计负责。(正确/错误)
\vspace{0.5em}\newline According to the Work Safety Law, the designers and design units of safety facilities for construction projects shall be responsible for the design of the safety facilities. (True/False)
\vspace{0.5em}\newline\textbf{Answer}: 正确 (True)
% \vspace{0.5em}\newline\textbf{Explanation}: /
&
\vspace*{-7ex}
\textbf{Question}: 安全生产“十三五”规划指出，广泛开展面向群众的安全教育活动，推动安全知识、安全。(正确/错误)
\vspace{0.5em}\newline The 13th Five-Year Plan for Work Safety states that safety education activities targeting the general public should be widely carried out to promote safety knowledge and safety. (True/False)
\vspace{0.5em}\newline\textbf{Answer}: 正确 (True)
% \vspace{0.5em}\newline\textbf{Explanation}: /
\\
\bottomrule
\end{tabularx}
\caption{Examples of multiple-choice, single-choice, and true/false questions in the Production Safety domain.}

\label{tab:PS}
\end{table*}

\end{CJK}

\begin{CJK}{UTF8}{gbsn}

\begin{table*}[ht]
\centering
\tiny
\renewcommand{\arraystretch}{1.2}  % optional: controls line spacing within cells
\begin{tabularx}{\textwidth}{
  >{\centering\arraybackslash}m{2cm}|
  >{\raggedright\arraybackslash}X|
  >{\raggedright\arraybackslash}X}
\toprule
\textbf{Category} & \makecell[c]{\textbf{Example 1}} & \makecell[c]{\textbf{Example 2}} \\
\midrule

\rule{0pt}{22ex}\shortstack{Oil and Gas\\(Multiple Choices)}
&
\vspace*{-12ex}
\textbf{Question}: 城镇燃气企业的安全操作规程要明确各部门、各岗位工作流程的()。
\vspace{0.5em}\newline The safety operation procedures of urban gas enterprises shall specify the following aspects of the workflow of each department and each position: ()
\vspace{0.5em}\newline A．工作量
\newline B．衔接关键点
\newline C．安全管理点
\newline D．绩效标准
\vspace{0.5em}\newline A. Workload
\newline B. Key connection points
\newline C. Safety management points
\newline D. Performance standards
\vspace{0.5em}\newline\textbf{Answer}: BC
% \vspace{0.5em}\newline\textbf{Explanation}: /
&
\vspace*{-12ex}
\textbf{Question}: 火灾逃生策略的“三救”中包括()。
\vspace{0.5em}\newline The “three rescue” strategies for fire escape include: ()
\vspace{0.5em}\newline A．结伴同行互“救”
\newline B．选择最近的电梯自“救”
\newline C．结绳下滑“救”
\newline D．向外界求“救”
\vspace{0.5em}\newline A. Accompanying each other and mutually “rescuing”
\newline B. Choosing the nearest elevator for self-“rescue”
\newline C. Rope-assisted sliding “rescue”
\newline D. Calling for “rescue” from the outside
\vspace{0.5em}\newline\textbf{Answer}: CD
% \vspace{0.5em}\newline\textbf{Explanation}: /
\\
\midrule
\rule{0pt}{24ex}\shortstack{Oil and Gas\\(Single Choice)}
&
\vspace*{-13ex}
\textbf{Question}: 我国()个人从事管道燃气经营活动。
\vspace{0.5em}\newline China () individuals engaging in piped gas business operations.
\vspace{0.5em}\newline A．鼓励
\newline B．允许
\newline C．禁止
\newline D．指导
\vspace{0.5em}\newline A. Encourages
\newline B. Permits
\newline C. Prohibits
\newline D. Guides
\vspace{0.5em}\newline\textbf{Answer}: C
% \vspace{0.5em}\newline\textbf{Explanation}: /
&
\vspace*{-13ex}
\textbf{Question}: 燃气经营者停业应在()前向所在地燃气管理部门报告，经批准方可停业。
\vspace{0.5em}\newline A gas operator intending to suspend business operations shall report to the local gas administration department () in advance, and may only suspend operations upon approval.
\vspace{0.5em}\newline A．60日
\newline B．60个工作日
\newline C．90日
\newline D．90个工作日
\vspace{0.5em}\newline A. 60 calendar days
\newline B. 60 working days
\newline C. 90 calendar days
\newline D. 90 working days
\vspace{0.5em}\newline\textbf{Answer}: D
% \vspace{0.5em}\newline\textbf{Explanation}: /
\\
\midrule
\rule{0pt}{12ex}\shortstack{Oil and Gas\\(True False)}
&
\vspace*{-7ex}
\textbf{Question}: 已建危险化学品生产装置需要停产的，由本级人民政府决定并组织实施。(正确/错误)
\vspace{0.5em}\newline If an existing hazardous chemical production facility needs to suspend production, the decision and organization for implementation shall be made by the people’s government at the same administrative level. (True/False)
\vspace{0.5em}\newline\textbf{Answer}: 正确 (True)
% \vspace{0.5em}\newline\textbf{Explanation}: /
&
\vspace*{-7ex}
\textbf{Question}: 《危险化学品重大危险源监督管理暂行规定》不适用城镇燃气的安全监督管理。(正确/错误)
\vspace{0.5em}\newline The Interim Provisions on the Supervision and Administration of Major Hazard Installations of Hazardous Chemicals do not apply to the safety supervision and administration of urban gas. (True/False)
\vspace{0.5em}\newline\textbf{Answer}: 正确 (True)
% \vspace{0.5em}\newline\textbf{Explanation}: /
\\
\bottomrule
\end{tabularx}
\caption{Examples of multiple-choice, single-choice, and true/false questions in the Oil and Gas domain.}
\label{tab:OG}
\end{table*}

\end{CJK}

\begin{CJK}{UTF8}{gbsn}

\begin{table*}[ht]
\centering
\tiny
\renewcommand{\arraystretch}{1.2}  % optional: controls line spacing within cells
\begin{tabularx}{\textwidth}{
  >{\centering\arraybackslash}m{2cm}|
  >{\raggedright\arraybackslash}X|
  >{\raggedright\arraybackslash}X}
\toprule
\textbf{Category} & \makecell[c]{\textbf{Example 1}} & \makecell[c]{\textbf{Example 2}} \\
\midrule

\rule{0pt}{26ex}\shortstack{Fire Safety\\(Multiple Choices)}
&
\vspace*{-14ex}
\textbf{Question}: 《消防救援队伍作战训练安全行动手册》是根据()等法律规范，结合灭火救援作战安全工作实际制定。
\vspace{0.5em}\newline The Operational Training Safety Action Manual for Fire and Rescue Teams was formulated based on () and other legal norms, in combination with the actual safety practices of firefighting and rescue operations.
\vspace{0.5em}\newline A．《中华人民共和国消防法》
\newline B．《执勤战斗条令》
\newline C．《消防员职业健康标准》
\newline D．《宪法》
\vspace{0.5em}\newline A. Fire Protection Law of the People’s Republic of China
\newline B. Regulations on Duty Combat Operations
\newline C. Occupational Health Standards for Firefighters
\newline D. The Constitution
\vspace{0.5em}\newline\textbf{Answer}: ABC
% \vspace{0.5em}\newline\textbf{Explanation}: /
&
\vspace*{-14ex}
\textbf{Question}: 各级消防救援队伍党政主要负责同志为本级作战训练安全工作第一责任人，()为灭火救援作战训练安全直接责任人。
\vspace{0.5em}\newline The principal Party and government leaders at all levels of the fire and rescue teams are the primary persons responsible for operational training safety at their respective levels. () are the direct persons responsible for safety in firefighting and rescue operational training.
\vspace{0.5em}\newline A．大队长
\newline B．分管领导
\newline C．业务部门领导
\newline D．现场指挥员(训练组织者)
\vspace{0.5em}\newline A. Battalion chief
\newline B. Leader in charge
\newline C. Head of the functional department
\newline D. On-site commander (training organizer)
\vspace{0.5em}\newline\textbf{Answer}: BCD
% \vspace{0.5em}\newline\textbf{Explanation}: /
\\
\midrule
\rule{0pt}{30ex}\shortstack{Fire Safety\\(Single Choice)}
&
\vspace*{-16ex}
\textbf{Question}: 以我国消防队伍配备的某型躯(肢)体固定气囊为例，其技术性能参数表述错误的是()。
\vspace{0.5em}\newline Taking as an example a certain type of body (limb) immobilization airbag equipped by China's fire rescue teams, which of the following descriptions of its technical performance parameters is incorrect? ()
\vspace{0.5em}\newline A、PVC材料制成，表面不容易损坏，可洗涤。
\newline B、可保持形状60h以上。
\newline C、可按伤员的各种形态而变化。
\newline D、用X光、CT、MRI检查时可穿透。
\vspace{0.5em}\newline A. Made of PVC material, surface is not easily damaged, washable.
\newline B. Can maintain its shape for more than 60 hours.
\newline C. Can adapt to various postures of the injured person.
\newline D. Can be penetrated during X-ray, CT, and MRI examinations.
\vspace{0.5em}\newline\textbf{Answer}: B
% \vspace{0.5em}\newline\textbf{Explanation}: /
&
\vspace*{-16ex}
\textbf{Question}: 《队列条令》规定，国家综合性消防救援队伍人员必须严格执行本条令，加强队列训练，培养良好的姿态、严整的队容、过硬的作风、严格的纪律性和协调一致的动作，促进队伍()建设，巩固和提高战斗力。
\vspace{0.5em}\newline According to the Formation Regulations, personnel of the national comprehensive fire and rescue teams must strictly implement these regulations, strengthen formation training, cultivate proper posture, neat appearance, strong work style, strict discipline, and coordinated actions, in order to promote the construction of () within the team, and to consolidate and enhance combat effectiveness.
\vspace{0.5em}\newline A．正规化
\newline B．军事化
\newline C．规范化
\newline D．整齐化
\vspace{0.5em}\newline A. Regularization
\newline B. Militarization
\newline C. Standardization
\newline D. Orderliness
\vspace{0.5em}\newline\textbf{Answer}: A
% \vspace{0.5em}\newline\textbf{Explanation}: /
\\
\midrule
\rule{0pt}{15ex}\shortstack{Fire Safety\\(True False)}
&
\vspace*{-8ex}
\textbf{Question}: 评选先进基层单位要用是否能够完成重大任务来衡量。(正确/错误)
\vspace{0.5em}\newline The selection of exemplary grassroots units should be measured by whether they are able to accomplish major tasks. (True/False)
\vspace{0.5em}\newline\textbf{Answer}: 错误 (False)
% \vspace{0.5em}\newline\textbf{Explanation}: /
&
\vspace*{-8ex}
\textbf{Question}: 作战行动应根据指挥员指令，编组实施，至少三人以上协同配合，同进同出，严禁擅自行动。(正确/错误)
\vspace{0.5em}\newline Operational actions shall be carried out according to the commander’s instructions, organized in groups, with at least three persons cooperating in coordination, entering and exiting together, and unauthorized actions are strictly prohibited. (True/False)
\vspace{0.5em}\newline\textbf{Answer}: 错误 (False)
% \vspace{0.5em}\newline\textbf{Explanation}: /
\\
\bottomrule
\end{tabularx}
\caption{Examples of multiple-choice, single-choice, and true/false questions in the Fire Safety domain.}
\label{tab:FS}
\end{table*}

\end{CJK}

\begin{CJK}{UTF8}{gbsn}

\begin{table*}[ht]
\centering
\tiny
\renewcommand{\arraystretch}{1.2}  % optional: controls line spacing within cells
\begin{tabularx}{\textwidth}{
  >{\centering\arraybackslash}m{2cm}|
  >{\raggedright\arraybackslash}X|
  >{\raggedright\arraybackslash}X}
\toprule
\textbf{Category} & \makecell[c]{\textbf{Example 1}} & \makecell[c]{\textbf{Example 2}} \\
\midrule

\rule{0pt}{37ex}\shortstack{Civil Engineering\\(Multiple Choices)}
&
\vspace*{-19ex}
\textbf{Question}: 公路建设必须执行国家环境保护和资源节约的法律法规，应作环境影响评价和水土保持方案评价的包括()。
\vspace{0.5em}\newline Highway construction must comply with national laws and regulations on environmental protection and resource conservation. Projects for which environmental impact assessment and soil and water conservation plan evaluation must be conducted include: ()
\vspace{0.5em}\newline A．高速公路
\newline B．一、二级公路
\newline C．三级公路
\newline D．有特殊要求的公路建设项目
\vspace{0.5em}\newline A. Expressways
\newline B. Class I and II highways
\newline C. Class III highways
\newline D. Highway construction projects with special requirements
\vspace{0.5em}\newline\textbf{Answer}: ABD
\vspace{0.5em}\newline\textbf{Explanation}: 根据《公路建设项目环境影响评价规范》(JTGB03-2006)第1.0.3条规定，本规范适用于需编制报告书的新建或改扩建的高速公路、一级公路和二级公路建设项目的环境影响评价，其他等级的公路建设项目环境影响评价可参照执行。
\vspace{0.5em}\newline According to Clause 1.0.3 of the Specifications for Environmental Impact Assessment of Highway Construction Projects (JTGB03-2006), this specification applies to environmental impact assessments that require the preparation of full reports for new or expanded/renovated expressways, Class I highways, and Class II highway construction projects. Environmental impact assessments for highway projects of other grades may be carried out by reference to this standard.
&
\vspace*{-19ex}
\textbf{Question}: 通信设施应提供哪些信息服务平台？()
\vspace{0.5em}\newline Which information service platforms should be provided by communication facilities? ()
\vspace{0.5em}\newline A．语音
\newline B．数据
\newline C．图像
\newline D．控制信号
\vspace{0.5em}\newline A. Voice
\newline B. Data
\newline C. Image
\newline D. Control signals
\vspace{0.5em}\newline\textbf{Answer}: ABC
\vspace{0.5em}\newline\textbf{Explanation}: 根据《高速公路交通工程及沿线设施设计通用规范》(JTGD80-2006)第7.5.1条规定，通信设施应根据高速公路通信网络规划，统一技术标准，统一进网要求，保证已建和在建高速公路通信系统的互联互通；通信系统应为用路者和管理者提供语音、数据、图像信息交互服务宽带网络平台。
\vspace{0.5em}\newline According to Clause 7.5.1 of the General Specifications for Highway Traffic Engineering and Roadside Facilities Design (JTGD80-2006), communication facilities shall comply with the highway communication network plan, adopt unified technical standards and network access requirements, and ensure interconnectivity between existing and under-construction highway communication systems. The communication system shall provide a broadband network platform for voice, data, and image information interaction services to both road users and managers.
\\
\midrule
\rule{0pt}{40ex}\shortstack{Civil Engineering\\(Single Choice)}
&
\vspace*{-21ex}
\textbf{Question}: 根据《民法典》，执行政府定价或政府指导价的合同时，对于逾期交付标的物的处置方式是()。
\vspace{0.5em}\newline According to the Civil Code, for contracts subject to government pricing or government-guided pricing, what is the treatment method for delayed delivery of the subject matter? ()
\vspace{0.5em}\newline A. 遇价格上涨时，按照原价格执行；价格下降时，按照新价格执行
\newline B. 遇价格上涨时，按照新价格执行；价格下降时，按照原价格执行
\newline C. 无论价格上涨或下降，均按照新价格执行
\newline D. 无论价格上涨或下降，均按照原价格执行
\vspace{0.5em}\newline A. In case of a price increase, the original price applies; in case of a price decrease, the new price applies
\newline B. In case of a price increase, the new price applies; in case of a price decrease, the original price applies
\newline C. Regardless of price increase or decrease, the new price applies
\newline D. Regardless of price increase or decrease, the original price applies
\vspace{0.5em}\newline\textbf{Answer}: A
\vspace{0.5em}\newline\textbf{Explanation}: 《民法典》规定，执行政府定价或政府指导价的，在合同约定的交付期限内政府价格调整时，按照交付时的价格计价。逾期交付标的物的，遇价格上涨时，按照原价格执行；价格下降时，按照新价格执行。逾期提取标的物或者逾期付款的，遇价格上涨时，按照新价格执行；价格下降时，按照原价格执行。
\vspace{0.5em}\newline The Civil Code stipulates that for contracts under government pricing or government-guided pricing, if the government price is adjusted within the agreed delivery period, the transaction shall be priced at the delivery-time price.
If the delivery of the subject matter is delayed, then in the case of a price increase, the original price applies; in the case of a price decrease, the new price applies. If the buyer delays collection of the subject matter or delays payment, then in the case of a price increase, the new price applies; in the case of a price decrease, the original price applies.
&
\vspace*{-21ex}
\textbf{Question}: 某工程招标估算价3000万元，根据《招标投标法实施条例》的规定，则投标保证金最高不得超过()。
\vspace{0.5em}\newline For a certain project with a tender estimated price of 30 million RMB, according to the Regulations for the Implementation of the Bidding Law, the maximum amount of the bid security shall not exceed: ()
\vspace{0.5em}\newline A. 20万元
\newline B. 60万元
\newline C. 80万元
\newline D. 100万元
\vspace{0.5em}\newline A. 200,000 RMB
\newline B. 600,000 RMB
\newline C. 800,000 RMB
\newline D. 1,000,000 RMB
\vspace{0.5em}\newline\textbf{Answer}: B
\vspace{0.5em}\newline\textbf{Explanation}: 知识点：招标投标法实施条例。如招标人在招标文件中要求投标人提交投标保证金，投标保证金不得超过招标项目估算价的2\%。
\vspace{0.5em}\newline Knowledge point: Regulations for the Implementation of the Bidding Law. If the tendering party requires the bidders to submit a bid security in the tender documents, the bid security must not exceed 2\% of the estimated price of the bidding project.
\\
\midrule
\rule{0pt}{27ex}\shortstack{Civil Engineering\\(True False)}
&
\vspace*{-14ex}
\textbf{Question}: 《建筑施工企业主要负责人、项目负责人和专职安全生产管理人员安全生产管理规定》（中华人民共和国住房和城乡建设部令第17号）第十七条规定，项目负责人对本项目安全生产管理全面负责，应当建立项目安全生产管理体系，明确项目管理人员安全职责，落实安全生产管理制度，确保项目安全生产费用有效使用。(正确/错误)
\vspace{0.5em}\newline According to Article 17 of the Regulations on the Safety Production Management of Principals, Project Managers, and Full-time Safety Managers of Construction Enterprises (Order No. 17 of the Ministry of Housing and Urban-Rural Development of the People's Republic of China), the project manager shall be fully responsible for the safety production management of the project, and shall establish a safety production management system for the project, clarify the safety responsibilities of project management personnel, implement the safety production management system, and ensure the effective use of safety production expenses for the project. (True/False)
\vspace{0.5em}\newline\textbf{Answer}: 正确 (True)
% \vspace{0.5em}\newline\textbf{Explanation}: /
&
\vspace*{-14ex}
\textbf{Question}: 《危险性较大的分部分项工程安全管理规定》第十条规定：实行施工总承包的，专项施工方案应当由施工总承包单位组织编制。(正确/错误)
\vspace{0.5em}\newline According to Article 10 of the Regulations on Safety Management of Sub-projects with High Risk, in the case of general contracting for construction, the special construction plan shall be organized and prepared by the general contractor. (True/False)
\vspace{0.5em}\newline\textbf{Answer}: 正确 (True)
% \vspace{0.5em}\newline\textbf{Explanation}: /
\\
\bottomrule
\end{tabularx}
\caption{Examples of multiple-choice, single-choice, and true/false questions in the Civil Engineering domain.}
\label{tab:CE}
\end{table*}

\end{CJK}

\begin{CJK}{UTF8}{gbsn}

\begin{table*}[ht]
\centering
\tiny
\renewcommand{\arraystretch}{1.2}  % optional: controls line spacing within cells
\begin{tabularx}{\textwidth}{
  >{\centering\arraybackslash}m{2cm}|
  >{\raggedright\arraybackslash}X|
  >{\raggedright\arraybackslash}X}
\toprule
\textbf{Category} & \makecell[c]{\textbf{Example 1}} & \makecell[c]{\textbf{Example 2}} \\
\midrule

\rule{0pt}{50ex}\shortstack{Economics and Finance\\(Multiple Choices)}
&
\vspace*{-26ex}
\textbf{Question}: 当某公司决定裁减部分员工时，其做法错误的是()。
\vspace{0.5em}\newline When a company decides to lay off some employees, which of the following practices is incorrect? ()
\vspace{0.5em}\newline A．裁员人数未达到职工总人数的10\%，可以随时实施裁员
\newline B．裁减人员在20人以上的，应当向当地劳动行政部门报告裁减人员方案，批准后方可裁员
\newline C．裁减人员未达到20人的，不用向劳动行政部门报告裁减人员方案
\newline D．应当在裁减人员前15日向工会全体职工说明情况，听取工会或职工的意见
\newline E．裁减人员时应考虑优先留用的相关人员
\vspace{0.5em}\newline A. If the number of layoffs does not reach 10\% of the total number of employees, the company may implement the layoffs at any time
\newline B. If 20 or more employees are to be laid off, the layoff plan must be reported to the local labor administration department and may only be implemented after approval
\newline C. If fewer than 20 employees are to be laid off, there is no need to report the layoff plan to the labor administration department
\newline D. The company shall explain the situation to the trade union and all employees 15 days prior to the layoff and solicit their opinions
\newline E. When laying off employees, the company shall consider retaining employees with priority qualifications
\vspace{0.5em}\newline\textbf{Answer}: ABCD
\vspace{0.5em}\newline\textbf{Explanation}: 《劳动合同法》第四十一条规定，用人单位因实施裁员解除劳动合同。有下列情形之一，裁减人员20人以上或者裁减不足20人但占企业职工总数10\%以上的，用人单位提前30日向工会或全体职工说明情况，听取工会或者职工意见后，裁减人员方案经向劳动行政部门报告，可以裁减人员。裁减人员方案只需要报告劳动行政部门即可，不需要等待劳动行政部门的批复同意。
\vspace{0.5em}\newline According to Article 41 of the Labor Contract Law, if an employer terminates labor contracts due to staff reductions, and one of the following conditions is met - laying off 20 or more employees, or laying off fewer than 20 employees but amounting to more than 10\% of the company’s total workforce - the employer must explain the situation to the trade union or all employees 30 days in advance, solicit opinions, and report the layoff plan to the labor administration department. The layoff plan only needs to be reported to the labor authority; approval is not required.
&
\vspace*{-26ex}
\textbf{Question}: 根据《社会保险法》规定，保险关系可以随本人转移的有()。
\vspace{0.5em}\newline According to the Social Insurance Law, which of the following types of insurance relationships can be transferred along with the individual? ()
\vspace{0.5em}\newline A．基本养老保险
\newline B．基本医疗保险
\newline C．工伤保险
\newline D．生育保险
\newline E．企业年金
\vspace{0.5em}\newline A. Basic pension insurance
\newline B. Basic medical insurance
\newline C. Work-related injury insurance
\newline D. Maternity insurance
\newline E. Enterprise annuity
\vspace{0.5em}\newline\textbf{Answer}: AB
\vspace{0.5em}\newline\textbf{Explanation}: 根据《社会保险法》规定，保险关系可以随本人转移的有基本养老保险、基本医疗保险、失业保险。企业年金的个人缴费虽然可以转移，但它不属于社会保险范畴。
\vspace{0.5em}\newline According to the Social Insurance Law, the insurance relationships that can be transferred along with the individual include basic pension insurance, basic medical insurance, and unemployment insurance. Although personal contributions to enterprise annuities can be transferred, enterprise annuities do not fall within the scope of social insurance.
\\
\midrule
\rule{0pt}{35ex}\shortstack{Economics and Finance\\(Single Choice)}
&
\vspace*{-18ex}
\textbf{Question}: 根据公司法，重要的国有独资公司合并、分离、解散，应当由()批准。
\vspace{0.5em}\newline According to the Company Law, the merger, division, or dissolution of an important wholly state-owned enterprise shall be approved by: ()
\vspace{0.5em}\newline A. 上级人民政府
\newline B. 本级国资监管机构
\newline C. 本级人民政府
\newline D. 上级国资监管机构
\vspace{0.5em}\newline A. The higher-level people’s government
\newline B. The state-owned assets supervision and administration authority at the same level
\newline C. The people’s government at the same level
\newline D. The higher-level state-owned assets supervision and administration authority
\vspace{0.5em}\newline\textbf{Answer}: C
\vspace{0.5em}\newline\textbf{Explanation}: 本题考查国有独资公司的权力机构。重要的国有独资公司合并、分立、解散、申请破产的，应当由国有资产监督管理机构审核后，报本级人民政府批准。
\vspace{0.5em}\newline This question examines the authority over wholly state-owned enterprises. The merger, division, dissolution, or bankruptcy filing of an important wholly state-owned enterprise shall be reviewed by the state-owned assets supervision and administration authority, and submitted to the people’s government at the same level for approval.
&
\vspace*{-18ex}
\textbf{Question}: 劳动者可以随时单方面解除劳动合同的情形是()。
\vspace{0.5em}\newline In which of the following situations may an employee unilaterally terminate the labor contract at any time? ()
\vspace{0.5em}\newline A．劳动者在试期内的
\newline B．劳动者以欺诈、胁迫的手段或者乘人之危，使用人单位在违背真实意思的情况下订立劳动合同
\newline C．用人单位未给劳动者缴纳社会保险费的
\newline D．劳动者处于孕期、产期或哺乳期的
\vspace{0.5em}\newline A. The employee is within the probation period
\newline B. The employer concluded the labor contract by means of fraud, coercion, or taking advantage of the employee’s difficulties, thereby violating the employee’s true intention
\newline C. The employer fails to pay social insurance premiums for the employee
\newline D. The employee is in the pregnancy, maternity, or breastfeeding period
\vspace{0.5em}\newline\textbf{Answer}: C
\vspace{0.5em}\newline\textbf{Explanation}: 根据《劳动合同法》第三十八条规定，用人单位存在下列情况之一的，劳动者可以无须通知用人单位，单方面解除劳动合同，选项C符合。
\vspace{0.5em}\newline According to Article 38 of the Labor Contract Law, if the employer commits any of the following acts, the employee may unilaterally terminate the labor contract without notifying the employer. Option C meets this condition.
\\
\midrule
\rule{0pt}{15ex}\shortstack{Economics and Finance\\(True False)}
&
\vspace*{-8ex}
\textbf{Question}: 发行人民币、管理人民币流通”是《中华人民共和国中国人民银行法》赋予中央银行的法定职责。(正确/错误)
\vspace{0.5em}\newline Issuing the renminbi and managing its circulation” is a statutory responsibility assigned to the central bank by the Law of the People's Republic of China on the People's Bank of China. (True/False)
\vspace{0.5em}\newline\textbf{Answer}: 正确 (True)
% \vspace{0.5em}\newline\textbf{Explanation}: /
&
\vspace*{-8ex}
\textbf{Question}: 金融机构为逃避人民银行的反假货币执法检查，销毁有关证据材料，人民银行将给予50-200万元的罚款处理。(正确/错误)
\vspace{0.5em}\newline If a financial institution, in order to evade the anti-counterfeit currency enforcement inspection by the People’s Bank of China, destroys relevant evidentiary materials, the People’s Bank of China shall impose a fine ranging from 500,000 to 2,000,000 RMB. (True/False)
\vspace{0.5em}\newline\textbf{Answer}: 错误 (False)
% \vspace{0.5em}\newline\textbf{Explanation}: /
\\
\bottomrule
\end{tabularx}
\caption{Examples of multiple-choice, single-choice, and true/false questions in the Economics and Finance domain.}
\label{tab:EF}
\end{table*}

\end{CJK}

\begin{CJK}{UTF8}{gbsn}

\begin{table*}[ht]
\centering
\tiny
\renewcommand{\arraystretch}{1.2}  % optional: controls line spacing within cells
\begin{tabularx}{\textwidth}{
  >{\centering\arraybackslash}m{2cm}|
  >{\raggedright\arraybackslash}X|
  >{\raggedright\arraybackslash}X}
\toprule
\textbf{Category} & \makecell[c]{\textbf{Example 1}} & \makecell[c]{\textbf{Example 2}} \\
\midrule

\rule{0pt}{34ex}\shortstack{Banking and Insurance\\(Multiple Choices)}
&
\vspace*{-18ex}
\textbf{Question}: 目前，我国中央银行的创新型货币政策工具包括()。
\vspace{0.5em}\newline At present, the innovative monetary policy tools of China's central bank include: ()
\vspace{0.5em}\newline A. 短期流动性调节工具
\newline B. 常备借贷便利
\newline C. 临时流动性便利
\newline D. 中期借贷便利
\newline E. 抵押补充贷款
\vspace{0.5em}\newline A. Short-term Liquidity Adjustment Tool
\newline B. Standing Lending Facility
\newline C. Temporary Liquidity Facility
\newline D. Medium-term Lending Facility
\newline E. Pledged Supplementary Lending
\vspace{0.5em}\newline\textbf{Answer}: ABCDE
\vspace{0.5em}\newline\textbf{Explanation}: 随着利率市场化改革的不断完善，我国中央银行创设了多种新型政策工具，包括短期流动性调节工具（SLO）、临时流动性便利（TLF）、常备借贷便利（SLF）、中期借贷便利（MLF）、抵押补充贷款（PSL），用以管理中短期利率水平。
\vspace{0.5em}\newline With the continuous improvement of interest rate liberalization reform, the central bank of China has introduced various innovative policy tools, including the Short-term Liquidity Adjustment Tool (SLO), Temporary Liquidity Facility (TLF), Standing Lending Facility (SLF), Medium-term Lending Facility (MLF), and Pledged Supplementary Lending (PSL), in order to manage medium- and short-term interest rates.
&
\vspace*{-18ex}
\textbf{Question}: 2013年1月16日以来，我国多层次股票市场包括()。
\vspace{0.5em}\newline Since January 16, 2013, China’s multi-tier stock market includes: ()
\vspace{0.5em}\newline A. 主板市场
\newline B. 全国中小企业股份转让系统
\newline C. 期货市场
\newline D. 中小企业板市场
\newline E. 创业板市场
\vspace{0.5em}\newline A. Main Board Market
\newline B. National Equities Exchange and Quotations
\newline C. Futures Market
\newline D. Small and Medium Enterprise Board Market
\newline E. Growth Enterprise Market
\vspace{0.5em}\newline\textbf{Answer}: ABDE
\vspace{0.5em}\newline\textbf{Explanation}: 我国多层次股票市场分为场内市场和场外市场，场内市场主要包括沪深主板市场、中小企业板市场和创业板市场，场外市场包括全国中小企业股份转让系统、区域股权交易市场以及已试点的券商柜台交易市场。
\vspace{0.5em}\newline China’s multi-tier stock market is divided into on-exchange and off-exchange markets. The on-exchange market mainly includes the Shanghai and Shenzhen Main Board Markets, the Small and Medium Enterprise Board Market, and the Growth Enterprise Market. The off-exchange market includes the National Equities Exchange and Quotations, regional equity trading markets, and pilot broker over-the-counter trading platforms.
\\
\midrule
\rule{0pt}{34ex}\shortstack{Banking and Insurance\\(Single Choice)}
&
\vspace*{-18ex}
\textbf{Question}: N股是我国股份公司在()上市的股票。
\vspace{0.5em}\newline N-shares refer to the stocks of Chinese joint-stock companies listed on ().
\vspace{0.5em}\newline A. 新加坡
\newline B. 纽约
\newline C. 香港
\newline D. 伦敦
\vspace{0.5em}\newline A. Singapore
\newline B. New York
\newline C. Hong Kong
\newline D. London
\vspace{0.5em}\newline\textbf{Answer}: B
\vspace{0.5em}\newline\textbf{Explanation}: N股是指由中国境内注册的公司发行、直接在美国纽约上市的股票。
\vspace{0.5em}\newline N-shares refer to stocks issued by companies registered in mainland China that are directly listed on the New York Stock Exchange in the United States.
&
\vspace*{-18ex}
\textbf{Question}: 目前，我国实行以市场供求为基础、参考()。
\vspace{0.5em}\newline At present, China implements an exchange rate regime based on market supply and demand, with reference to ().
\vspace{0.5em}\newline A．美元和欧元进行调节、自由浮动的汇率制度
\newline B．美元进行调节、可调整的盯住汇率制度
\newline C．欧元进行调节、可调整的盯住汇率制度
\newline D．一篮子货币进行调节、有管理的浮动汇率制度
\vspace{0.5em}\newline A. U.S. dollar and Euro for adjustment, a freely floating exchange rate regime
\newline B. U.S. dollar for adjustment, an adjustable pegged exchange rate regime
\newline C. Euro for adjustment, an adjustable pegged exchange rate regime
\newline D. A basket of currencies for adjustment, a managed floating exchange rate regime
\vspace{0.5em}\newline\textbf{Answer}: D
\vspace{0.5em}\newline\textbf{Explanation}: 汇率制度又称汇率安排，是指一国货币当局对其货币汇率的变动所作的一系列安排或规定的统称。目前，我国实行以市场供求为基础、参考一篮子货币进行调节、有管理的浮动汇率制度。
\vspace{0.5em}\newline The exchange rate regime, also known as exchange rate arrangement, refers to a set of arrangements or regulations made by a country's monetary authority regarding changes in its currency’s exchange rate. Currently, China adopts a managed floating exchange rate regime based on market supply and demand, with reference to a basket of currencies for adjustment.
\\
\midrule
\rule{0pt}{22ex}\shortstack{Banking and Insurance\\(True False)}
&
\vspace*{-12ex}
\textbf{Question}: 在我国，贷款基准利率是指商业银行对其最优质客户执行的贷款利率，其他贷款利率可在此基础上加减点生成。(正确/错误)
\vspace{0.5em}\newline In China, the benchmark lending rate refers to the loan interest rate applied by commercial banks to their best-quality clients, and other lending rates can be generated by adding or subtracting basis points from this rate. (True/False)
\vspace{0.5em}\newline\textbf{Answer}: 错误 (False)
\vspace{0.5em}\newline\textbf{Explanation}: 中国人民银行对商业银行的再贷款利率，可以理解为我国目前的基准利率。贷款基础利率是指商业银行对其最优质客户执行的贷款利率，其他贷款利率可在此基础上加减点生成。
\vspace{0.5em}\newline The re-lending rate set by the People’s Bank of China for commercial banks can be considered as the current benchmark interest rate in China. The loan prime rate refers to the lending rate that commercial banks apply to their best-quality clients, and other loan interest rates can be formed by adding or subtracting basis points based on this rate.
&
\vspace*{-12ex}
\textbf{Question}: 我国的货币政策工具逐步从价格型向数量型转变。(正确/错误)
\vspace{0.5em}\newline China's monetary policy instruments are gradually shifting from price-based to quantity-based. (True/False)
\vspace{0.5em}\newline\textbf{Answer}: 错误 (False)
\vspace{0.5em}\newline\textbf{Explanation}: 近年来，随着宏观经济的变化，我国的货币政策工具逐步从数量型向价格型转变。
\vspace{0.5em}\newline In recent years, with changes in the macroeconomic environment, China’s monetary policy instruments have gradually shifted from quantity-based to price-based.
\\
\bottomrule
\end{tabularx}
\caption{Examples of multiple-choice, single-choice, and true/false questions in the Banking and Insurance domain.}
\label{tab:BI}
\end{table*}

\end{CJK}